\documentclass[10pt, a4paper]{article}
\usepackage{lrec2022} 
\usepackage{multibib}
\newcites{languageresource}{Language Resources}
\usepackage{graphicx}
\usepackage{tabularx}
\usepackage{soul}
\usepackage{amssymb}
\usepackage{amsmath}
\usepackage{multirow} 
\usepackage{multicol} 
\usepackage{arydshln}
\usepackage{url}
\usepackage{float}
\usepackage{booktabs}
\usepackage{amsfonts}

\usepackage{titlesec}
\titleformat{\section}{\normalfont\large\bfseries\center}{\thesection.}{1em}{}
\titleformat{\subsection}{\normalfont\SmallTitleFont\bfseries\raggedright}{\thesubsection.}{1em}{}
\titleformat{\subsubsection}{\normalfont\normalsize\bfseries\raggedright}{\thesubsubsection.}{1em}{}
\renewcommand\thesection{\arabic{section}}
\renewcommand\thesubsection{\thesection.\arabic{subsection}}
\renewcommand\thesubsubsection{\thesubsection.\arabic{subsubsection}}

\usepackage{epstopdf}
\usepackage[utf8]{inputenc}

\usepackage{hyperref}
\usepackage{xstring}

\usepackage{color}

\title{\textbf{The Uncertainty-based Retrieval Framework for}\\
Ancient Chinese CWS and POS}

\name{Pengyu Wang, Zhichen Ren} 

\address{Fudan University, Tongji University \\
         220 Handan Road Shanghai, China\\ 4800 Caoan Highway, Shanghai, China \\
         wpyjihuai@gmail.com, 1850091@tongji.edu.cn}

\abstract{
Automatic analysis for modern Chinese has greatly improved the accuracy of text mining in related fields, but the study of ancient Chinese is still relatively rare. Ancient text division and lexical annotation are important parts of classical literature comprehension, and previous studies have tried to construct auxiliary dictionary and other fused knowledge to improve the performance. In this paper, we propose a framework for ancient Chinese Word Segmentation and Part-of-Speech Tagging that makes a twofold effort: on the one hand, we try to capture the wordhood semantics; on the other hand, we re-predict the uncertain samples of baseline model by introducing external knowledge. The performance of our architecture outperforms pre-trained BERT with CRF and existing tools such as Jiayan.
 \\ \newline  }

\begin{document}

\maketitleabstract

\section{Introduction}

Chinese Word Segmentation (CWS) and Part-of-Speech (POS) Tagging are two important tasks of natural language processing. With the rapid development of deep learning and pre-trained models, the performance of CWS and POS Tagging increased significantly. A simple model using pre-trained  BERT and conditional random field (CRF) can reach a high accuracy. Since words are the most common components in a Chinese sentence and words can cause ambiguity, structures that can capture word information have been used in these tasks to get better performance. 

Lexicon-based methods have been widely used in CWS, Chinese POS tagging and NER tasks to capture wordhood information \cite{yang2018subword,li2020flat}. These methods can leverage semantic information of words and improve model performance. However, lexicon-based methods have several drawbacks. One of the most severe problems is that they depend heavily on the quality of lexicons. Unfortunately, building an ancient Chinese lexicon is more difficult than building a modern Chinese lexicon, since there are few ancient Chinese corpus, and words from different corpus are different. 

Further, sentences in ancient Chinese are always shorter than sentences in Chinese, which means words in ancient Chinese have a richer meaning and can cause misunderstanding or wrong classification. 

The two problems mentioned above make ancient Chinese CWS and POS Tagging a harder problem. In our model, we combine bigram features with BERT to capture wordhood information in sentences. The semantic information of bigram plays a similar role to the lexicon, while it is unnecessary to build a large lexicon for ancient Chinese corpus. To deal with the ambiguity, or uncertainty in sentences, we use MC-dropout method to find uncertain parts of sentences. Next we use a Knowledge Fusion Model to retrieve auxillary knowledge and re-predict the uncentain parts. Our experiments show that our model outperforms pre-trained BERT model \url{https://huggingface.co/SIKU-BERT/sikuroberta} with CRF and Jiayan \url{https://github.com/jiaeyan/Jiayan} in our dataset \textit{Zuozhuan}.

\section{Background and Related Work}
\par
\subsection{CWS and POS Tagging}
Chinese Word Segmentation (CWS) is the fundamental of Chinese natural language understanding. It splits a sentence into several words, which are basic components of a Chinese sentence. CWS is necessary because there is no natural segmentation between Chinese words. Part-of-Speech Tagging (POS Tagging) further assigns POS tags for each word in a sentence.

\subsection{Knowledge Retrieval}
Knowledge retrieval is a method used to enhance the performance of language models, and they are most commonly used in NER tasks. Knowledge databases \cite{qiu2014automatic,gu2018search} and search engines \cite{geng2022turner} are used to retrieve knowledge, and the knowledge retrieved is used to argument the input sentences.

\section{Approach}

As previous work \cite{qiu2019multi,ke2020unified}, the CWS and POS Tagging task is viewed as a character-based sequence labeling problem. Specifically, given input sequence $X=[c_1,c_2,...,c_n]$ composed of continuous characters, the model should output a label sequence $Y=[y_1,y_2,...y_n]$ with  $y_i\ {\in}\ TagSet$.

In this section, we will introduce the improvement proposed for local semantic information capture, followed by the uncertainty sampling method. Finally, we will introduce our overall framework utilizing the uncertainty sampling method. 

\subsection{Local Semantic Enhancement}

\begin{figure}[!h]
\centering

\includegraphics[width=\columnwidth]{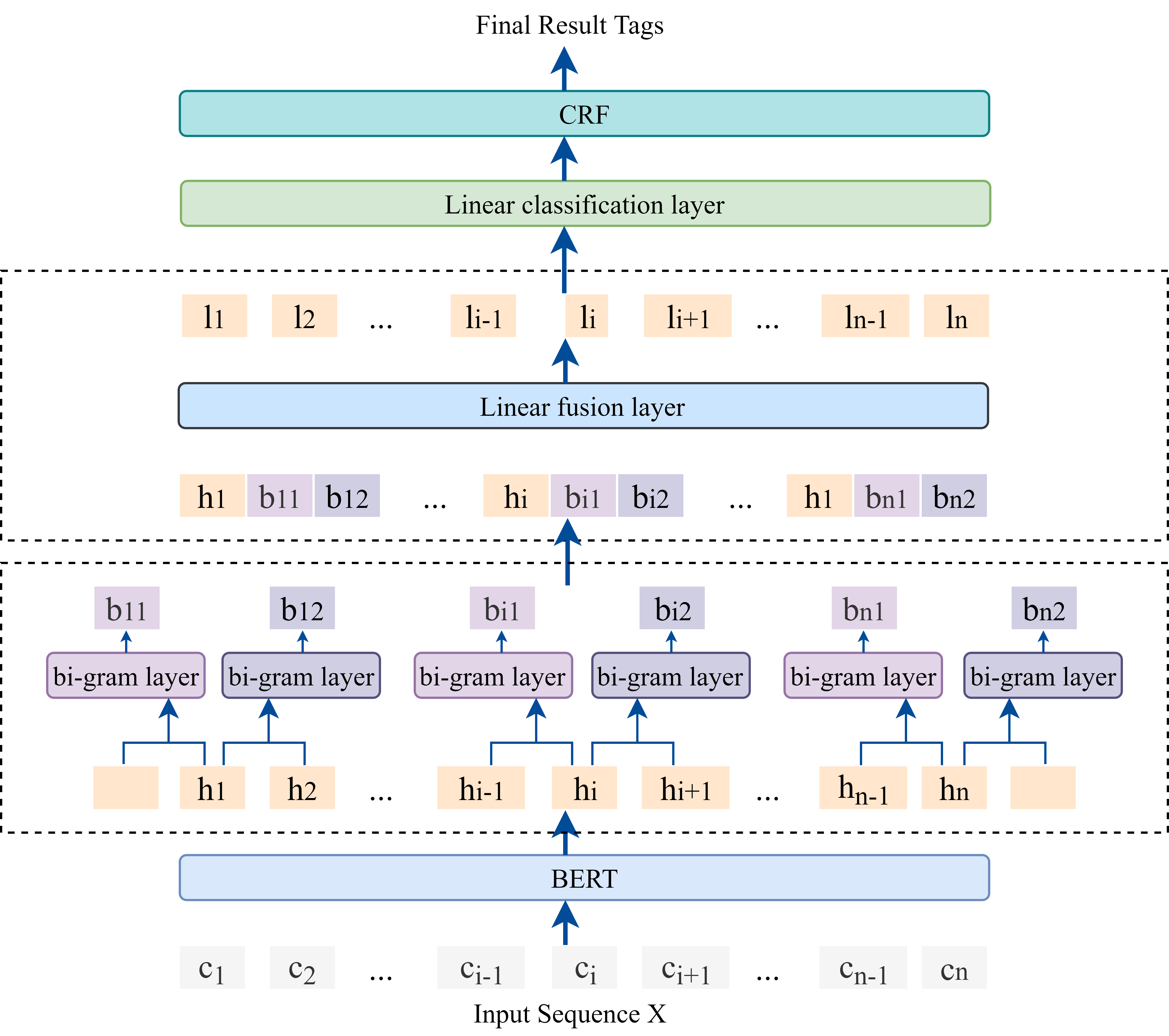} 

\caption{Architecture of baseline model.}
\label{semantic_enhancement}
\end{figure}

BERT \cite{devlin2018bert} is a Transformer based bidirectional language model, which solves the problem of long-term dependence in RNN models. However, this also makes BERT lose the ability to capture local semantic features. Therefore, we integrated the bigram features to introduce local semantic information. The overall architecture of our baseline model is displayed in Figure \ref{semantic_enhancement}, and we call it \textit{Semantic Enhancement BERT}.

\subsubsection{Encoder}
Given input sequence $X=[c_1,c_2,...,c_n], X\ {\in}\ \mathbb{R}^n$. We employ BERT as our basic encoder, converting X to hidden character representations as follows,

\begin{equation}
    H=BERT(X),
\end{equation}
where $H\ {\in}\ \mathbb{R}^{n{\times}d_h}$.
\subsubsection{Linear Bigram Layer}
The vocabulary of ancient Chinese is short, consise and meaningful, and the bigram features have proved beneficial for CWS \cite{chen2017adversarial,ke2020unified}. Therefore, we construct the bigram concatenated vectors for every character $c_i$ by concatenating it's hidden character representations with the previous character's and the latter character's. Then we convert the concatenated vectors to bigram feature vectors $b_{i1}$, $b_{i2}$ by two Linear bigram layer as follows, 

\begin{align}
    &b_{i1}=LinearLayer_1(h_{i-1}{\oplus}h_i),\\
    &b_{i2}=LinearLayer_2(h_{i}{\oplus}h_{i+1}),
\end{align}

where $b_{i1},b_{i2}\ {\in}\ \mathbb{R}^{d_b}$.

\subsubsection{Linear Fusion Layer}

We construct Composite feature vector $h_i$ for character $c_i$ by concatenating $h_i$, $b_{i1}$ and $b_{i2}$ as follows,

\begin{equation}
h_i^{'}=h_i\ {\oplus}\ b_{i1}\ {\oplus}\ b_{i2},
\end{equation}
where $h_i^{'}\ {\in}\ \mathbb{R}^{(d_h+2{\times}d_b)}$.

$H'$ is defined as follows,
\begin{equation}
    H^{'}=[h_1^{'},h_2^{'},...,h_n^{'}].
\end{equation}

Then, we use a simple fusion mechanism to convert the Composite feature vectors to Fusion feature vectors by a Linear Layer,

\begin{equation}
    L=LinearLayer_3(H^{'}),
\end{equation}
where $H^{'}\ {\in}\ \mathbb{R}^{n{\times}(d_h+2{\times}d_b)},L\ {\in}\ \mathbb{R}^{n{\times}d_1}$.

\subsubsection{Decoder}
The Fusion feature representations are converted into the probabilities over the POS labels by an MLP layer,

\begin{align}
    P^T&=Softmax(WL^T+b),
\end{align}
where $P\ {\in}\ \mathbb{R}^{n{\times}d_t}$. $d_t$ is the number of POS tags. $P_{ik}$ represents the probability that the label of $c_i$ is $tag_k$.

Finally, we decode $P$ using \textbf{Viterbi algorithm} to obtain the final tag sequence $T=[t_1,t_2,...,t_n],T\ {\in}\ \mathbb{R}^n$.

\subsection{Uncertainty Sampling}
BERT is already very powerful. Under the condition that the annotated dataset is very limited, simply increasing the complexity of the model structure will not make performance better. So we introduce uncertainty sampling and knowledge retrieving.

\subsubsection{Uncertainty Sampling Method}
MC Dropout \cite{gal2016dropout} is a general approach to obtain the uncertain components. Formally, given input sequence $X$, we first obtain the provisional label sequence $T_p$ utilizing trained baseline model. Then, we utilize MC dropout to keep dropout active and generate $k$ candidate label sequences $T_1,T_2,...,T_k$ with Viterbi decoding. 
The difference between each candidate-predicted word set and the provisional-predicted word set can be considered uncertain words. Then we obtain uncertain components by merging all overlapping uncertain words.

\subsubsection{Preliminary Statistics}
Similar to \newcite{geng2022turner}'s evaluation approach, we conduct an investigation on test set of two Ancient Chinese datasets to verify the importance of the uncertainty component. We use \textit{Semantic Enhancement BERT} as baseline model and generate 8 candidate label sequences using MC dropout. The results are displayed in Table 3.

\begin{table}[!h]
\centering
\begin{tabularx}{\columnwidth}{lll}

      \toprule
      &\textit{Zuozhuan}&\textit{Shiji}\\
      \midrule
      CWS F1 Score & 95.606\% & 93.465\%\\
      CWS Oracle F1 Score & 97.777\% & 96.780\%\\
      POS F1 Score & 91.229\% & 87.618\%\\
      POS Oracle F1 Score & 95.602\% & 93.417\%\\
      ACC$_{uncertain}$ & 57.190\% & 55.951\%\\
      ACC$_{certain}$ & 94.560\% & 91.704\%\\
      \bottomrule

\end{tabularx}
\caption{The statistics of the uncertain components. \textbf{F1 Score} denotes the F1 score of the baseline model on the test dataset. \textbf{Oracle F1 Score} denotes the F1 score obtained by the baseline model if the labels of the uncertain components are corrected. \textbf{ACC$_{uncertain}$} and \textbf{ACC$_{certain}$} denote the label accuracy of the provisional results for the uncertain components and the confident components, respectively.}
\label{mcresult}
\end{table}

The significant gap between certain components and uncertain components indicates that the uncertain components are real hard components and become bottlenecks for performance. Therefore, by querying about uncertain components, the ancient corpus with the same structure can be retrieved. 

\subsubsection{Retrieving}
Different from the retrieval idea in the NER task \cite{geng2022turner}, we first collect several Pre-Qin ancient texts to form our knowledge corpus. For word $w$ corresponding to each uncertain component, we query the sentences containing $w$. In particular, if the uncertain component contains only one character, we construct bigram words $w_1$ and $w_2$ for the character $w$ by concatenating it with the previous character and the latter character. Then we look for sentences containing $w_1$ or $w_2$ instead of $w$. 

We rank sentences by similarities in order to obtain sentences with grammatical structures similar to $X$. Generally, the similarity between two sentences $P$ and $Q$ is defined as follows,

\begin{equation}
    \mathrm{s}=\frac{\operatorname{union}(\mathrm{P}, \mathrm{Q})}{\|\mathrm{P}\|+\|\mathrm{Q}\|},
    \label{similarity}
\end{equation}

where $union(P,Q)$ is the total number of the same characters in $P$ and $Q$, $\|\mathrm{P}\|$ and $\|\mathrm{Q}\|$ is the length of $P$ and $Q$, respectively. Finally, we choose the most similar sentences as auxiliary knowledge.

\subsection{Framework}

\begin{figure}[!h]
\centering

\includegraphics[width=\columnwidth]{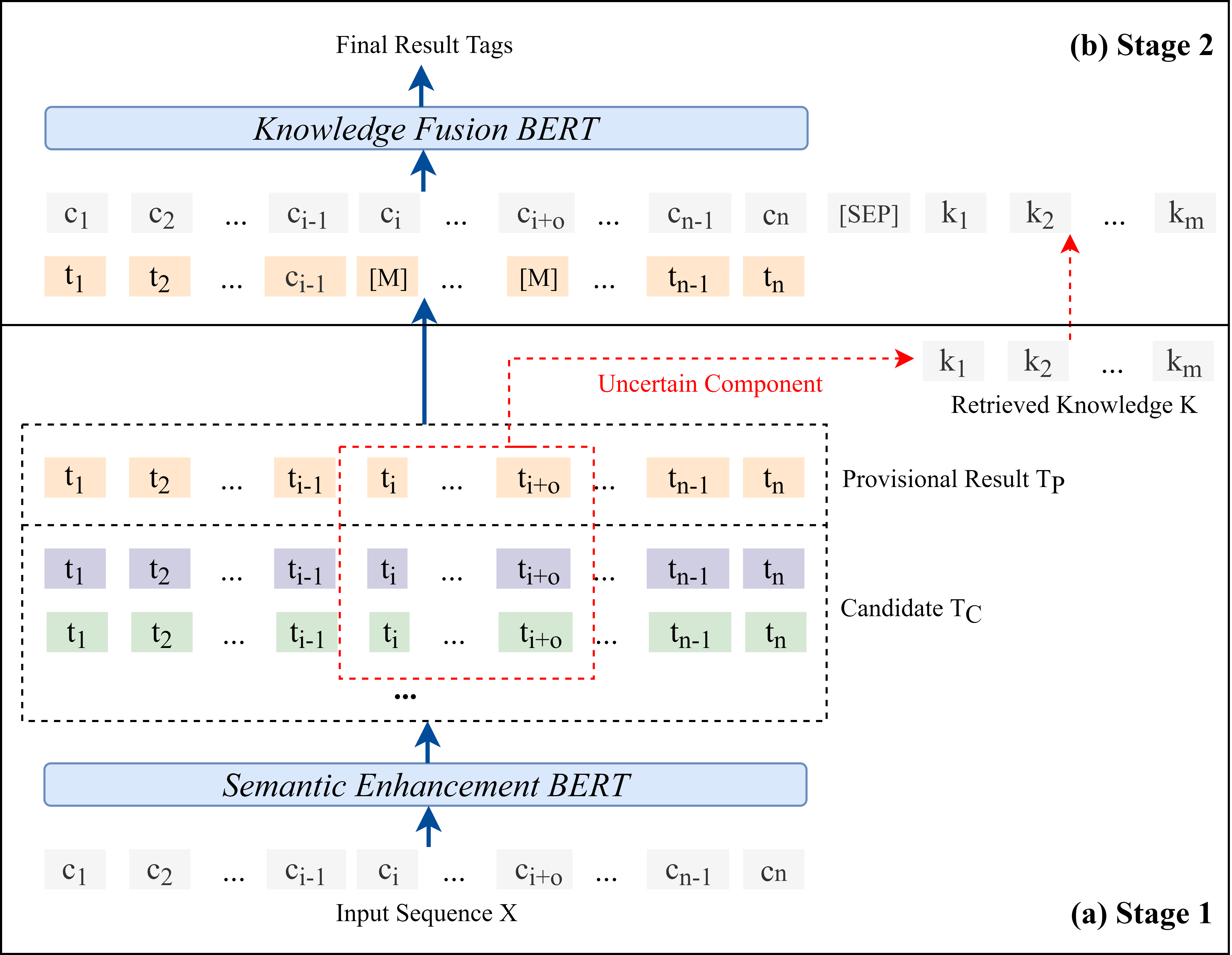} 

\caption{The overall framework.}
\label{overall}
\end{figure}

In this part, we will present our overall framework, which is displayed in Figure \ref{overall}.

\subsubsection{Stage One: Provisional Results and Uncertainty Sampling}
Given input sequence $X=[c_1,c_2,...,c_n]$, we employ baseline model to obtain the provisional label sequence $T_p$ and candidate label sequences. Then we obtain the uncertain component $U=[c_i,c_{i+1},...c_{i+o}]$ using the method in Section 3.2.

If $X$ has no uncertain component, $T_p$ will be taken as the final prediction label sequence $T$. Otherwise, we use $U$ to retrieve the auxiliary knowledge $K$. If there are multiple uncertain components, we retrieve them separately and process them independently using the method in Stage Two.

\begin{table*}[!ht]  
	\centering
	\begin{tabular}{ccccc}
		\toprule
		\multirow{2}*{Model} & \multicolumn{2}{c}{Test-\textit{Zuozhuan}} &  \multicolumn{2}{c}{Test-\textit{Shiji}} \\ 
		& CWS-F1(\%) & POS-F1(\%) & CWS-F1(\%) & POS-F1(\%)\\
		\hline 
		Jiayan & 82.022 & / & 83.141 & / \\
        \midrule
		Siku-RoBERTa + CRF  & 96.073 & 91.998 & 92.937 & 87.466 \\
        \midrule
		SE-BERT  & 96.018 & 92.019 & 93.092 & 87.594 \\
		SE-BERT$^+$  & 96.148 & 92.292 & 92.914 & 86.691 \\
		{\bf SE-BERT$^+$+KF-BERT}  & {\bf 96.284} & {\bf 92.410} & {\bf 93.596} & {\bf 87.873} \\
        \midrule
		BERT-Bigram & 96.009 & 91.853 & 93.015 & 87.574 \\
		\bottomrule
	\end{tabular}
	\caption{Jiayan is an NLP toolkit focusing on ancient Chinese processing. SE-BERT denotes \textit{Semantic Enhancement BERT} using Siku-RoBERTa, SE-BERT$^+$ denotes \textit{Semantic Enhancement BERT} using Siku-RoBERTa$^+$ as pre-trained BERT, and KF-BERT means \textit{Knowledge Fusion BERT} using Siku-RoBERTa. BERT-Bigram denotes Siku-RoBERTa incorporating pre-trained bigram embedding. To utilize the entire training set, we use cross-validation and average the prediction results of K models, where K = 5.
}
\label{table:performance}
\end{table*}

\subsubsection{Stage Two: Knowledge Fusion Prediction}
In the second stage, we re-predict the label sequence of input sequence $X$ by combining the auxiliary knowledge $K$ and the provisional label sequence $T_p$ obtained in Stage One.

Similar to \newcite{geng2022turner}, we concatenate $X$ and $K$ to obtain the knowledge-enhanced input sequence $X^{'}=[c_1,c_2,...,c_n,[SEP],k_1,k_2,...,k_m]$ and construct the auxilary label sequence as follows,

\begin{align}
    t_{i}^{\prime}&= \begin{cases}t_{i} & \text { if } i \leq n \text { and } c_{i} \notin U \\ {[M A S K]} & \text { if } c_{i} \in U \\ {[P A D]} & \text { if } i>n\end{cases},\\
    T^{'}&=[t_1^{'},t_2^{'},...,t_n^{'},t_{n+1}^{'},...,t_{n+m+1}^{'}].
\end{align}

Finally, we combine $X'$ and $T'$ as the input of Bert-based \textit{Knowledge Fusion BERT} (KF-BERT) to obtain the probability distribution $D$,

\begin{align}
    &E_{T^{'}}=LabelEmbedding(T^{'}),\\
    &E_{X^{'}}=CharacterEmbedding(X^{'}),\\
    &D\ \ =KF\mbox{-}BERT(E_{T^{'}}+E_{X^{'}}),
\end{align}

where $D=[d_1,d_2,...d_n]$ and $d_i$ is the probability distribution of $c_i$, and $d_{ij}$ is the probability of $c_i$ being predicted to $tag_j$.

\textit{Label Embedding} and \textit{Character Embedding} are parameters need to be trained. Finally, we get the final label sequence by \textbf{Viterbi algorithm}. In particular, if there are multiple uncertain components in X, we process them separately in the second stage and average all obtained $D$ before Viterbi decoding.

\section{Experiment}
We conducted a series of experiments to validate the effectiveness of our framework. We follow the competition EvaHan2022 \url{https://circse.github.io/LT4HALA/2022/EvaHan}, using a tag set containing 22 POS tags and a tag set \{B, M, E, S\} to denote the beginning, middle, and end of a word as well as single words. Thus we have a total of 88 tags for joint CWS and POS Tagging classification. We used the standard F1-Score as evaluation metric. All experiments were conducted on a server with 8 GeForce RTX 3090.

\subsection{Overall Performance}
Table \ref{table:performance} shows the over all performance and some ablation experiments.

From Table \ref{table:performance}, the performance of our model is much higher than the ancient Chinese processing toolkit Jiayan. our efforts in both semantic enhancement (siku-roBERTa+CRF and SE-BERT) and knowledge fusion (SE-BERT$^+$ and SE-BERT$^+$+KF-BERT) show that large improvements were achieved. 
We have also collected a lot of ancient Chinese text data, and further pretrained a BERT model (SE-BERT$^+$) \footnote{More details can be seen in \url{https://github.com/Jihuai-wpy/bert-ancient-chinese}.}.
SE-BERT$^+$ can further improve the performance. Our final model combines all the advantages and achieves good results.

\section{Discussion}
Regarding the combination of bigram features, we did not introduce new knowledge or more complex structures in our framework. \newcite{ke2020unified} incorporated pre-trained bigram embedding into their model. Referring to the work of \newcite{ke2020unified}, we conducted another experiment.

The experiment result in Table \ref{table:performance} shows that \textit{Semantic Enhancement BERT} works better than $BERT\mbox{-}Bigram$. However, the idea still shows a good direction for future research. The ancient vocabulary is short and rich in meaning, and the performance may be further improved if well pre-trained N-gram embedding can be properly introduced.

\section{Conclusion}
In this paper, we propose a framework for ancient Chinese CWS and POS Tagging that implements semantic enhancement and knowledge fusion. By utilizing bigram features and re-predicting the uncertain samples by fusing knowledge, our framework makes good predictions.

\section{References}\label{reference}

\bibliographystyle{lrec2022-bib}
\bibliography{lrec2022-example}

\begin{thebibliography}{}

\bibitem[\protect\citename{Chen \bgroup et al.\egroup }2017]{chen2017adversarial}
Chen, X., Shi, Z., Qiu, X., and Huang, X.
\newblock (2017).
\newblock Adversarial multi-criteria learning for chinese word segmentation.
\newblock {\em arXiv preprint arXiv:1704.07556}.

\bibitem[\protect\citename{Devlin \bgroup et al.\egroup }2018]{devlin2018bert}
Devlin, J., Chang, M.-W., Lee, K., and Toutanova, K.
\newblock (2018).
\newblock Bert: Pre-training of deep bidirectional transformers for language understanding.
\newblock {\em arXiv preprint arXiv:1810.04805}.

\bibitem[\protect\citename{Gal and Ghahramani}2016]{gal2016dropout}
Gal, Y. and Ghahramani, Z.
\newblock (2016).
\newblock Dropout as a bayesian approximation: Representing model uncertainty in deep learning.
\newblock In {\em international conference on machine learning}, pages 1050--1059. PMLR.

\bibitem[\protect\citename{Geng \bgroup et al.\egroup }2022]{geng2022turner}
Geng, Z., Yan, H., Yin, Z., An, C., and Qiu, X.
\newblock (2022).
\newblock Turner: The uncertainty-based retrieval framework for chinese ner.
\newblock {\em arXiv preprint arXiv:2202.09022}.

\bibitem[\protect\citename{Gu \bgroup et al.\egroup }2018]{gu2018search}
Gu, J., Wang, Y., Cho, K., and Li, V.~O.
\newblock (2018).
\newblock Search engine guided neural machine translation.
\newblock In {\em Proceedings of the AAAI Conference on Artificial Intelligence}, volume~32.

\bibitem[\protect\citename{Ke \bgroup et al.\egroup }2020]{ke2020unified}
Ke, Z., Shi, L., Meng, E., Wang, B., Qiu, X., and Huang, X.
\newblock (2020).
\newblock Unified multi-criteria chinese word segmentation with bert.
\newblock {\em arXiv preprint arXiv:2004.05808}.

\bibitem[\protect\citename{Li \bgroup et al.\egroup }2020]{li2020flat}
Li, X., Yan, H., Qiu, X., and Huang, X.
\newblock (2020).
\newblock Flat: Chinese ner using flat-lattice transformer.
\newblock {\em arXiv preprint arXiv:2004.11795}.

\bibitem[\protect\citename{Qiu \bgroup et al.\egroup }2014]{qiu2014automatic}
Qiu, X., Huang, C., and Huang, X.-J.
\newblock (2014).
\newblock Automatic corpus expansion for chinese word segmentation by exploiting the redundancy of web information.
\newblock In {\em Proceedings of COLING 2014, the 25th International Conference on Computational Linguistics: Technical Papers}, pages 1154--1164.

\bibitem[\protect\citename{Qiu \bgroup et al.\egroup }2019]{qiu2019multi}
Qiu, X., Pei, H., Yan, H., and Huang, X.
\newblock (2019).
\newblock Multi-criteria chinese word segmentation with transformer.
\newblock {\em arXiv preprint arXiv:1906.12035}.

\bibitem[\protect\citename{Yang \bgroup et al.\egroup }2018]{yang2018subword}
Yang, J., Zhang, Y., and Liang, S.
\newblock (2018).
\newblock Subword encoding in lattice lstm for chinese word segmentation.
\newblock {\em arXiv preprint arXiv:1810.12594}.

\end{thebibliography}

\section*{Appendix: Datasets and Hyperparameters}

The training and test datasets for this experiment are from the competition EvaHan2022 \url{https://circse.github.io/LT4HALA/2022/EvaHan}. The training and testa datasets were excerpted from \textit{Zuozhuan} and the testb dataset was excerpted from the \textit{Shiji}. The statistical information of the datasets is shown in Table \ref{stats}
\begin{table}[h]
\centering
\begin{tabularx}{0.8\columnwidth}{lll}

      \toprule
      &Size&Length$_{avg}$\\
      \midrule
      Train-\textit{Zuozhuan} & 1083K & 22.415\\
      Test-\textit{Zuozhuan} & 185K & 20.902\\
      Test-\textit{Shiji} & 352K & 29.302\\
      \bottomrule

\end{tabularx}
\caption{Dataset statistics.}
\label{stats}
\end{table}


\begin{table}[h]
\centering
\begin{tabularx}{\columnwidth}{lll}

      \hline
      &SE-Bert & KF-Bert\\
      \hline
      Epochs & 20 & 20\\
      Batch Size & 32 & 32\\
      Weight Decay & 0.1 & 0.1\\
      Dropout & 0.1 & 0.1\\
      Learning Rate & 1e-5 & 1e-5\\
      Optimizer & AdamW & AdamW\\
      Warm Up Ratio & 0.1 & 0.1\\
      Max Seq\_Len & 128 & 128\\
      $\alpha$ & - & \{0.1,1\}\\
      \hline

\end{tabularx}
\caption{Hyper parameters for \textit{Semantic Enhancement Bert} and \textit{Knowledge Fusion BERT}. }
\label{hyperparameters}
\end{table}

The hyper parameters are listed in table \ref{hyperparameters}. 

To enhance the learning of uncertain component, we introduce weight coefficient ${\omega}_i$ to set different weights for uncertain components and certain components so that the model pays more attention to the prediction of uncertain parts. The loss function L is defined as Eq. \eqref{loss},

\begin{align}
    &L=\frac{\sum_{\mathrm{i}}^{1 \leq \mathrm{i} \leq \mathrm{n}} \omega_{\mathrm{i}} \cdot \operatorname{loss}_{\mathrm{i}}}{\sum_{\mathrm{i}}^{1 \leq \mathrm{i} \leq \mathrm{n}} \omega_{\mathrm{i}}},
    \label{loss}
\end{align}
\begin{align}
    &\omega_{\mathrm{i}}= \begin{cases}1 & \text { if } c_{i} \in U \\ \alpha & \text { if } c_{i} \notin U\end{cases},
\end{align}
where ${\omega}_i$ is the weight coefficient at position $i$. $loss_i$ is the cross-entropy loss at position $i$. $\alpha$ is a hyper parameter ranges in $[0,1]$. In particular, we do not make predictions for auxiliary knowledge, nor do we calculate the loss of this part.

\end{document}